\documentclass[11pt]{article}
\usepackage[utf8]{inputenc}
\usepackage[T1]{fontenc}

\usepackage{amsmath,amssymb,amsthm}
\usepackage{booktabs}
\usepackage{graphicx}
\usepackage{caption}
\usepackage{multirow}
\usepackage{url}
 \usepackage{natbib}
 \usepackage[margin=1in]{geometry}
 
\theoremstyle{definition}
\newtheorem{definition}{Definition}[section]
\newtheorem{implication}{Implication}[section]

\newcommand{\E}{\mathbb{E}}

\newcommand{\sphere}{\mathbb{S}}

\title{A Geometric Taxonomy of Hallucination in LLMs}

\author{
  Javier Marín \\
  \textit{javier@jmarin.info}
}

\date{May 2026}

\begin{document}
\maketitle

\begin{abstract}
Hallucinations in deployed language models can have real consequences for downstream decisions in domains such as healthcare, legal, and financial services. In production, detection has to run on what the deployed system can see: the query, the response, and often a source document. White-box access to model internals and multi-sample querying are not generally available behind a third-party API. Within this setting --- black-box, single-pass, only question/answer available --- the dominant baseline is NLI, which returns a value but no diagnosis when it fails. We argue that operating directly on the geometry of the embedding space provides detection methods whose successes \emph{and failures} are interpretable as structural properties of contrastive sentence-encoder training \citep{wang2020understanding}. The contribution is: \emph{given an operationally-motivated taxonomy, geometry predicts which types of hallucination are detectable and which are not --- and the predictions hold.}

We propose three operational types organized by the relation of the response embedding to the plausibility region of grounded responses on the unit hypersphere, and derive from the alignment objective a prediction for each: (1)~query-proximate unfaithfulness is detectable by an angular ratio; (2)~confabulation outside the plausibility region produces a directional signature that outperforms NLI on expert-annotated error; (3)~factual errors sharing vocabulary and frame with correct answers are not separable by angular geometry. To validate on content resembling deployment, we built a 212-pair human-confabulated dataset across nine domains using provoked confabulation.
\end{abstract}

\section{Introduction}
\label{sec:introduction}

Large Language Models in production can hallucinate (produce outputs that are factually incorrect). This paper asks whether a deployed system can detect such outputs before they are acted on, and equally important, which outputs it \emph{cannot} detect. The deployment context narrows what detection can use. Production systems running on third-party APIs do not see model internals; multi-sample querying is impractical at scale; source-grounded methods require documents that are not always available. What remains is single-pass detection on the query and the response. Our standard baseline is entailment classification, usually implemented as NLI.

The term Hallucination covers different phenomena \citep{maynez2020faithfulness,ji2023survey,huang2025survey,zhang2023siren,hicks2024chatgpt,alansari2025comprehensive}. Recent geometric methods \citep{korun2026clusters,mir2025lsd,phillips2025geometric,catak2024uncertainty} operate at different levels of analysis (token clusters, hidden-state trajectories, sampled response polytopes) and offer empirically promising results. This paper adds an account, anchored in the alignment objective of contrastive sentence-encoder training \citep{wang2020understanding}, of which kinds of hallucination are geometrically detectable from displacement structure on the unit sphere and which are not. The alignment objective pulls semantically related pairs together as a function of distributional similarity, not truth value: two responses sharing vocabulary, topic, and syntactic frame will cluster on the sphere whether they are correct or not. This single property determines which failures leave geometric signatures and which do not.

Most hallucination benchmarks generate false content by prompting an LLM to produce false answers \citep{li2023halueval,ravi2024lynx,bang2025hallulens}. The resulting text carries traces of having been written under that instruction. A method's performance on such benchmarks therefore conflates two signals: the prompting trace and the actual hallucination. An alternative reference comes from human non-experts confabulating from memory: asked questions they cannot answer from knowledge, they write down whatever comes to mind, so the false content comes from incomplete memory rather than from instruction.

Three predictions organize the rest of the paper. The first is that query-proximate unfaithfulness leaves a measurable angular signature when context is available (Section~\ref{sec:type1}). The second is that Type~II confabulation --- content displaced outside the query's plausibility region --- produces a learnable directional signature (Section~\ref{sec:type2}). The third is that factual errors within correct conceptual frames are not separable from correct responses by angular geometry (Section~\ref{sec:type3}) --- and this third claim, if it holds, names a class of failures that geometric detection cannot reach. 

\section{Related Work}
\label{sec:related}

\subsection{Hallucination benchmarks}
Hallucination benchmarks differ in how false content is produced. Most use LLMs to generate wrong responses: HaluEval \citep{li2023halueval} uses ChatGPT to generate 35{,}000 hallucinated samples; HaluBench \citep{ravi2024lynx} and HalluLens \citep{bang2025hallulens} use similar LLM-generation protocols. Alternative resources --- FactScore \citep{min2023factscore}, RAGTruth \citep{niu2024ragtruth}, FaithBench \citep{bao2025faithbench} --- evaluate natural model outputs but do not provide paired controlled examples. TruthfulQA \citep{lin2022truthfulqa} targets shared misconceptions rather than domain-specific confabulation.

\subsection{Detection methods}
Existing detection methods differ in what they require to operate. Approaches that access model internals achieve strong performance: \citet{azaria2023internal} report AUROC 0.96 from hidden-state trajectories; \citet{chen2024inside} use eigenvalue spectra of response covariance matrices. These require white-box access unavailable in commercial API deployments. Multi-sample consistency methods sample the model repeatedly: SelfCheckGPT \citep{manakul2023selfcheckgpt} reaches AUC-PR 0.93 and semantic entropy \citep{farquhar2024detecting,kuhn2023semantic} reaches AUROC 0.79, but both require 5--20 generations per query. Source-grounded alignment methods perform well when context is available: \citet{vectara2024hhem} achieves 74--77\% balanced accuracy; AlignScore \citep{zha2023alignscore} reaches AUC-ROC 0.87. Our approach has three main features: black-box, single-pass, no source documents.

\subsection{Geometric detection}
Several methods developed within the past year propose to detect hallucinations based on geometric properties of embedding space. They differ from each other and from this work in the level of analysis, the access model, and the theoretical basis. \citet{korun2026clusters} introduces a three-type taxonomy based on token embedding cluster structure inside transformer encoders and decoders. \citet{mir2025lsd} (LSD) tracks the evolution of hidden-state representations across layers of the generating transformer; the method requires white-box access. \citet{phillips2025geometric} apply archetypal analysis to batches of responses sampled at non-zero temperature; the method is sampling-based. \citet{catak2024uncertainty} use convex-hull analysis on response embeddings. \citet{huang2026semantic} introduces a geodesic constraint on hidden-state trajectories during training; the constraint operates on the same geometric primitive --- displacement vectors between embeddings at different sequence positions --- but for an orthogonal purpose, enforcing trajectory linearity during training rather than measuring directional alignment at evaluation.

\section{A Taxonomy of Hallucinations}
\label{sec:taxonomy}

\subsection{Definitions}
We identify three types of hallucinations using an operational criterion: the relation of the response to the \emph{plausibility region} of the query. Throughout, $\phi : \mathcal{X} \to \mathbb{R}^d$ denotes a contrastively trained sentence encoder mapping text strings $\mathcal{X}$ to $d$-dimensional embeddings; $\hat\phi(s) = \phi(s) / \|\phi(s)\|_2$ denotes the $L_2$-normalized embedding lying on the unit hypersphere $\sphere^{d-1}$; and $\theta(a, b) = \arccos(\hat\phi(a)^\top \hat\phi(b)) \in [0, \pi]$ denotes the geodesic distance between two text embeddings on $\sphere^{d-1}$ \citep{docarmo1992riemannian, bridson1999metric}.

\begin{definition}[Plausibility region]
\label{def:plaus}

For a query $q$, the plausibility region $\mathcal{P}_q \subset \sphere^{d-1}$ is the set of embeddings $\hat\phi(r)$ of responses $r \in \mathcal{X}$ that are topically appropriate to $q$, where topically appropriate means responses sharing vocabulary, named entities, and syntactic frame with grounded responses to $q$ --- and that would therefore be alignment-related under contrastive training in the sense of \citet{wang2020understanding}.
\end{definition}

\begin{definition}[Type~I: Unfaithfulness, query-proximate variant]
\label{def:typeI}
Given a query $q \in \mathcal{X}$ and provided context $c \in \mathcal{X}$, a response $r \in \mathcal{X}$ is Type~I in the query-proximate sense if its embedding remains angularly closer on $\sphere^{d-1}$ to the embedding of $q$ than to the embedding of $c$: $\theta(r, q) < \theta(r, c)$, equivalently $\mathrm{SGI}(r; q, c) < 1$, where $\mathrm{SGI}$ is the Semantic Grounding Index defined in Section~\ref{sec:methods}.
\end{definition}

Definition~\ref{def:typeI} captures only one variant of unfaithfulness in the broader sense of \citet{maynez2020faithfulness}: responses whose embeddings ignore $c$ \emph{and} default to the query's semantic neighborhood. Other modes (off-topic continuations, parametric memory drawn from a different domain) produce different geometry and lie outside the scope of SGI.

\begin{definition}[Type~II: Confabulation]
\label{def:typeII}
A response $r \in \mathcal{X}$ is Type~II if its embedding lies outside the plausibility region of $q$: $\hat\phi(r) \notin \mathcal{P}_q$. Examples include invented institutions, redefined technical terms, and fabricated mechanisms.
\end{definition}

\begin{definition}[Type~III: Factual error within frame]
\label{def:typeIII}
A response $r \in \mathcal{X}$ is Type~III if its embedding lies inside the plausibility region of $q$ --- $\hat\phi(r) \in \mathcal{P}_q$ --- and $r$ contains incorrect propositional content. The wrong response and the correct response share vocabulary, named entities, and syntactic frame.
\end{definition}

A single response can show more than one type. The taxonomy classifies the dominant geometric signature. WikiBio-style annotation \citep{manakul2023selfcheckgpt} is a salient case where Type~II and Type~III errors are pooled under a single ``inaccurate'' label, with empirical consequences discussed in Section~\ref{sec:experiments_external}. The closest taxonomic work is \citet{korun2026clusters}, which proposes a three-type taxonomy on token-embedding cluster topology rather than sentence-level displacement. This approach operates at a different level than ours and can be considered complementary rather than competing.

\section{Geometric Implications of the Alignment Objective}
\label{sec:implications}

Contrastive training optimizes angular relationships on $\sphere^{d-1}$ \citep{reimers2019sentence,gao2021simcse}; \citet{wang2020understanding} formalize this through alignment and uniformity, and \citet{ethayarajh2019contextual} shows that contextual representations are highly anisotropic. Our taxonomy allows three predictions on the unit sphere, using the notation introduced in Section~\ref{sec:taxonomy}.

\begin{implication}[Type~I]
\label{imp:1}
A grounded response $r_g$ engages with $c$ and lies closer to $c$ than to $q$ on the sphere: $\theta(r_g,q) > \theta(r_g,c)$. A query-proximate unfaithful response $r_u$ remains nearer to $q$: $\theta(r_u,q) < \theta(r_u,c)$. Hence $\mathrm{SGI}(r_g;q,c)>1>\mathrm{SGI}(r_u;q,c)$.
\end{implication}

The motivation comes from the alignment objective~\citep{wang2020understanding}: when $r$ draws content from $c$, alignment-related pairs $(c,r)$ are pulled together on the sphere and $\hat\phi(r)-\hat\phi(q)$ acquires a component pointing toward the region of $c$.

\begin{implication}[Type~II]
\label{imp:2}
For a reference set $\mathcal{R}=\{(q_i,r_i)\}_{i=1}^N$ of verified grounded pairs, define the von Mises--Fisher mean grounding direction \citep{banerjee2005vmf}
\[
\hat\mu = \mu/\|\mu\|, \quad \mu = \tfrac{1}{N}\sum_i \hat\delta(q_i,r_i),
\]
with $\hat\delta(q,r) = (\hat\phi(r)-\hat\phi(q))/\|\hat\phi(r)-\hat\phi(q)\|$. A grounded response satisfies $\hat\delta(q,r)^\top\hat\mu \approx \E_{\mathcal{R}}[\hat\delta_i^\top\hat\mu]$. A confabulated response, displacing toward content semantically foreign to $\mathcal{P}_q$, satisfies $\hat\delta(q,r)^\top\hat\mu \ll$ this expectation.
\end{implication}

The strength of Prediction~\ref{imp:2} depends on whether $\mathcal{R}$ is drawn from a population in which a single $\hat\mu$ exists. We address this empirically. Under annotation conditions where confabulation is a clear semantic departure, $\hat\mu$ exists and transfers across subject domains. Under conditions that capture LLM generation artifacts or that include Types~II and III, $\hat\mu$ is condition-specific and does not transfer.

\begin{implication}[Type~III]
\label{imp:3}
Within-frame factual substitutions are not separable from correct responses by the angular geometry of $(q, r)$ pairs alone. Under the alignment objective \citep{wang2020understanding}, two responses sharing vocabulary, named entities, and syntactic frame cluster on $\sphere^{d-1}$ regardless of truth value. As a concrete instance: for $q = $ What is the capital of Australia?'', the displacement $\hat\phi(\text{Sydney''}) - \hat\phi(q)$ is statistically indistinguishable from $\hat\phi(\text{``Canberra''})$ - $\hat\phi(q)$, because both responses share the syntactic frame, the topical register, and the named-entity scaffolding of the question, and the alignment objective pulls them into the same neighborhood of $\sphere^{d-1}$ regardless of which one is true.
\end{implication}

Prediction~\ref{imp:3} is a working hypothesis. It rests on the alignment objective being computed over distributional features that do not encode truth conditions. A formal proof for arbitrary contrastively-trained encoders is open. The TruthfulQA evidence in Section~\ref{sec:type3} provides empirical support and identifies what \emph{is} detectable in such datasets (annotation style), distinguishing this from positive factual-error signal.

\section{Methodology}
\label{sec:experiments}

\subsection{A protocol for human-confabulated datasets}
\label{sec:methodology}

The protocol uses provoked confabulation \citep{berlyne1972confabulation,moscovitch1995confabulation}: a person is asked a question they do not know the answer to and produces an answer anyway. The pipeline has four steps: (1) select a domain and a question whose answer admits a single canonical form; (2) generate a grounded response from authoritative sources and verify it; (3) have a non-expert produce a response from memory without consulting any source; (4) match response lengths within $\pm$20\% to prevent length itself from becoming a detection shortcut (a general concern in shortcut learning).

\subsubsection{Confabulation mechanisms}
\label{sec:confab_mechanisms}
Confabulation in clinical neuropsychology has long been observed to produce different patterns depending on the kind of knowledge being asked about --- knowledge about rules and procedures, knowledge structured by templates, and declarative knowledge about how the world works each fail in characteristic ways when memory is incomplete \citep{dallabarba1993different}. We observe the same three-way structure in the dataset, with each pattern carrying a different geometric consequence in embedding space.

\emph{Register preservation} (regulatory and procedural domains) inverts factual claims while keeping the domain register intact; under the distributional hypothesis the confabulation is an embedding-space neighbor of the grounded response. Consider a question about Roth IRA contribution rules: the confabulation fabricates a flat 15\% withdrawal tax and invents mandatory distributions at 65, but every term (\emph{contributions, distributions, adjusted gross income, surcharge}) stays in the financial-regulatory register. The alignment objective clusters text by co-occurrence statistics \citep{wang2020understanding}; since the wrong answer shares the same lexical distribution as the right one, $\|\hat\phi(r_{\text{confab}}) - \hat\phi(r_{\text{grounded}})\|$ is small and detection by displacement is geometrically constrained.

\emph{Template-filling} (technical specification domains) preserves the syntactic template (entity, classification, mechanism, interaction) while swapping content in the slots; the template dominates the embedding, the slot content where truth resides contributes little. Asked about SQL \texttt{GROUP BY}, the confabulator describes parallel query partitioning rather than row aggregation, but the clause-mechanism-usage template is identical. Under alignment, the template's distributional signature --- shared function words, clause structure, technical register --- accounts for most of the embedding variance; the factual slot content occupies a low-variance subspace, making its substitution nearly invisible to cosine-based methods.

\emph{Mechanism inversion} (declarative-knowledge domains) preserves the broad subject register but invents a mechanism drawn from adjacent knowledge, importing vocabulary from the adjacent sub-register; the accumulated vocabulary shift moves the response away from the grounded one in embedding space. Asked how CRISPR-Cas9 edits genes, the confabulator describes protein-folding correction rather than DNA cleavage, importing \emph{chaperones, mRNA, refolding} from a neighboring molecular-biology register. This is the most detectable mechanism because the imported vocabulary creates distributional divergence: the co-occurrence statistics of protein-folding terms differ measurably from those of gene-editing terms, producing a displacement $\hat\delta$ that deviates from $\hat\mu$.

\subsection{Human-generated benchmark}
We generate a 212-pair dataset across nine deployment-relevant domains: Python coding (47), finance (40), medical (40), science (21), TypeScript coding (18), history (14), law (11), general knowledge (11), and geography (10). The premise is: \emph{given this question, which kind of answer, if it went undetected, would propagate into a wrong downstream decision?} Mean response lengths are matched between grounded ($\mu=55.6$ words) and confabulated ($\mu=57.3$). The 212-pair validation comes from a single English-speaking non-expert; person-specific patterns are a limitation of this validation, not of the methodology, which is designed to be applied by multiple generators across languages and domains. Table~\ref{tab:dataset_appendix} reports the full composition.

\begin{table}[h]
\centering
\small
\caption{Domain composition of the 212-pair human-confabulated dataset.}
\label{tab:dataset_appendix}
\begin{tabular}{@{}lc@{}}
\toprule
\textbf{Domain} & $n$ \\
\midrule
Python coding & 47 \\
Finance & 40 \\
Medical & 40 \\
Science & 21 \\
TypeScript coding & 18 \\
History & 14 \\
Law & 11 \\
General knowledge & 11 \\
Geography & 10 \\
\midrule
\textbf{Total} & \textbf{212} \\
\bottomrule
\end{tabular}
\end{table}

\paragraph{Register preservation (Roth IRA).}
Asked about contribution rules and tax treatment for a Roth IRA, the confabulator may invert tax treatment from after-tax to pre-tax, fabricate employer matching rules that do not exist for Roth accounts, and invent every numerical limit. The vocabulary remains exclusively financial-regulatory --- \emph{contributions, distributions, deductions, surcharge} --- so the response sits as an embedding-space neighbor of the grounded answer despite every factual claim being wrong.

\paragraph{Template-filling (\texttt{GROUP BY} in SQL).}
Asked about the function of \texttt{GROUP BY}, the confabulator may describe it as parallel query partitioning across worker nodes rather than as row aggregation by column value. The clause-and-table register is identical to the correct answer; the syntactic template (clause, mechanism, when-to-use) is preserved; only the slot content describing what \texttt{GROUP BY} actually does is wrong.

\paragraph{Mechanism inversion (CRISPR-Cas9).}
Asked how CRISPR-Cas9 edits genes, the confabulator may describe protein-folding correction rather than DNA cleavage and repair --- importing the protein-folding sub-register (\emph{chaperones, mRNA, refolding}). The biological domain register is preserved at the high level, but the specific mechanism vocabulary belongs to a different molecular-biology sub-discipline, producing the accumulated vocabulary shift that makes this mechanism the most detectable of the three.

\subsection{Metrics}
\label{sec:methods}

\paragraph{Semantic Grounding Index.}
For a query $q$, provided context $c$, and response $r$, with $\theta(\cdot, \cdot)$ as introduced in Section~\ref{sec:taxonomy}, the Semantic Grounding Index can be defined as
\begin{equation}
\mathrm{SGI}(r; q, c) = \frac{\theta(r, q)}{\theta(r, c)},
\label{eq:sgi}
\end{equation}
for $\theta(r, c) > 0$. By the strict-positivity properties of the geodesic distance on $\sphere^{d-1}$, $\mathrm{SGI}(r; q, c) > 0$ for all valid inputs, and Prediction~\ref{imp:1} translates directly into the threshold form
\[
\mathrm{SGI}(r; q, c) > 1 \iff \theta(r, q) > \theta(r, c),
\]
i.e., $r$ is angularly closer to $c$ than to $q$ on $\sphere^{d-1}$. Grounded responses satisfy $\mathrm{SGI} > 1$; query-proximate unfaithful responses satisfy $\mathrm{SGI} \leq 1$.

\paragraph{Directional Grounding Index.}
For a reference set $\mathcal{R} = \{(q_i, r_i)\}_{i=1}^N$ of grounded pairs, we define the unit-norm displacement of a $(q, r)$ pair as
\begin{equation}
\hat\delta(q, r) = \frac{\hat\phi(r) - \hat\phi(q)}{\|\hat\phi(r) - \hat\phi(q)\|_2},
\label{eq:delta}
\end{equation}
for $\hat\phi(r) \neq \hat\phi(q)$. With $\hat\mu$ as in Prediction~\ref{imp:2}, the Directional Grounding Index is defined as
\begin{equation}
\Gamma(q, r; \mathcal{R}) = \hat\delta(q, r)^\top \hat\mu \in [-1, +1].
\label{eq:gamma}
\end{equation}
$\Gamma$ is the cosine of the angle between the displacement of the test pair and the mean grounding direction estimated from $\mathcal{R}$: $\Gamma = +1$ corresponds to perfect alignment with the grounded reference direction, $\Gamma = 0$ to orthogonality, and $\Gamma < 0$ to opposing direction. Prediction~\ref{imp:2} states that grounded responses concentrate near $\E_{\mathcal{R}}[\hat\delta_i^\top \hat\mu]$ while confabulated responses are displaced toward smaller values of $\Gamma$.

A local variant $\Gamma_k$ replaces $\hat\mu$ with a query-specific direction $\hat\mu_q$ estimated from the $k$ grounded pairs $(q_i, r_i) \in \mathcal{R}$ whose queries $q_i$ are nearest to $q$ on $\sphere^{d-1}$:
\begin{equation}
\hat\mu_q = \frac{\mu_q}{\|\mu_q\|_2}, \quad \mu_q = \frac{1}{k} \sum_{i \in \mathcal{N}_k(q)} \hat\delta(q_i, r_i),
\label{eq:mu_local}
\end{equation}
where $\mathcal{N}_k(q) = \{i : q_i\}$ is among the $k$ nearest neighbors of $q$ in $\{q_j\}_{j=1}^N$. Computing global $\Gamma$ requires one embedding and one dot product, $O(d)$ per query after one-time $O(Nd)$ precomputation of $\hat\mu$; computing $\Gamma_k$ requires $O(Nd)$ per query for the nearest-neighbor selection unless an index is precomputed.

\paragraph{Statistical analysis.}
We report AUROC with $1{,}000$-resample bootstrap 95\% CIs, permutation $p$-values ($5{,}000$ resamples), Hanley--McNeil $z$-tests \citep{hanley1982meaning}, Wilson 95\% CIs \citep{wilson1927probable} for proportions, and Fisher's exact tests for $2 \times 2$ comparisons.

\paragraph{Baselines.}
We compare $\Gamma$ to NLI (\texttt{cross-encoder/nli-deberta-v3-small}; \citealp{he2020deberta}) as the strongest single-pass black-box alternative on $(q,r)$ inputs, and to cosine similarity $\cos(\hat\phi(q), \hat\phi(r))$ as the simplest geometric baseline. We do not compare to white-box methods \citep{azaria2023internal,chen2024inside,mir2025lsd} ---different access model---, to multi-sample methods \citep{manakul2023selfcheckgpt,farquhar2024detecting,kuhn2023semantic,phillips2025geometric} ---different cost profile---, or to AlignScore \citep{zha2023alignscore} on context-free experiments ---it requires source documents.

\subsection{Embedding models}

We use five sentence encoders: \texttt{sentence-t5-large} \citep{ni2021sentencet5} (primary, 768d), \texttt{all-mpnet-base-v2} \citep{reimers2019sentence} (768d), \texttt{all-MiniLM-L6-v2} \citep{reimers2019sentence} (384d), \texttt{bge-small-en-v1.5} \citep{xiao2024cpack} (384d), and \texttt{gte-small} \citep{li2023gte} (384d).

\section{Results}

\subsection{Type~I: SGI on grounded context}
\label{sec:type1}

Table~\ref{tab:sgi_results} reports SGI on HaluEval QA ($n=10{,}000$, \citealp{li2023halueval}) across five embedding architectures. Mean SGI is 1.180 for grounded responses and 0.910 for hallucinated ones. The cross-model Pearson correlation of per-instance SGI scores averages $r=0.85$, confirming that SGI measures a property of the text rather than an artifact of any single embedding.

\begin{table}[h]
\centering
\small
\caption{SGI on HaluEval QA ($n=10{,}000$). Mean AUROC 0.805 across architectures. NLI and cosine baselines on the same data.}
\label{tab:sgi_results}
\begin{tabular}{@{}lccc@{}}
\toprule
\textbf{Method / Model} & \textbf{Grnd.} & \textbf{Halluc.} & \textbf{AUROC} \\
\midrule
sentence-t5-large & 1.203 & 0.856 & 0.824 \\
all-mpnet-base-v2 & 1.142 & 0.921 & 0.776 \\
bge-base-en-v1.5 & 1.231 & 0.948 & 0.823 \\
e5-base-v2 & 1.138 & 0.912 & 0.794 \\
all-MiniLM-L6-v2 & 1.188 & 0.913 & 0.806 \\
\cmidrule{1-4}
SGI Mean & 1.180 & 0.910 & 0.805 \\
\midrule
NLI (DeBERTa-v3-sm) & --- & --- & 0.748 \\
Cosine (sent-t5-lg) & --- & --- & 0.941 \\
\bottomrule
\end{tabular}
\end{table}

These results confirm Prediction~\ref{imp:1}: query-proximate unfaithfulness is detectable. Cosine similarity alone reaches 0.941 --- HaluEval hallucinations diverge far enough from correct answers that surface proximity suffices for separation. NLI (0.748) and SGI (0.805) are lower because they measure structural properties rather than proximity. But SGI provides what the others cannot: the ratio directly diagnoses \emph{whether} a response drew from context or defaulted to the query, which neither cosine distance nor NLI scores decompose. SGI does not detect unfaithfulness modes that displace away from both $q$ and $c$; for those, source-grounded methods such as AlignScore \citep{zha2023alignscore} are more appropriate.

\subsection{Type~II: three-way detection comparison}
\label{sec:type2}

We test whether detection performance measured on LLM-generated benchmarks transfers to human-confabulated content. Three conditions on the same 212 questions: human confabulations from Section~\ref{sec:methodology}; LLM confabulations generated by prompting an LLM (claude-sonnet-4-5 accessed via API in February 2026) to produce plausible false responses to the same questions; and a 500-pair random sample from HaluEval QA \citep{li2023halueval}. We establish as detection rate the fraction of pairs where $\cos(q,g) > \cos(q,c)$, with random baseline 50\%.

\begin{table*}[h]
\centering
\small
\caption{Three-way detection comparison across four embedding architectures. Det\%: fraction with $\cos(q,g)>\cos(q,c)$. CI: Wilson 95\%. $\Delta$: mean $\cos(q,g)-\cos(q,c)$. Paired: mean $\cos(g,c)$. $^{***}p<.001$ vs.\ Human (Wilcoxon signed-rank). All Human and LLM rates significantly above 50\% ($p<10^{-8}$, binomial). NLI on same 212-pair human confabulations: 57.5\% detection (AUROC 0.536).}
\label{tab:three_way}
\begin{tabular}{@{}llcccc@{}}
\toprule
\textbf{Model} & \textbf{Condition} & \textbf{Det\%} & \textbf{95\% CI} & $\Delta$ & \textbf{Paired} \\
\midrule
\multirow{3}{*}{all-MiniLM-L6-v2} & Human & 69.3\% & [62.8, 75.2] & $+.044$ & .722 \\
& LLM & 72.6\% & [66.3, 78.2] & $+.027$ & .858 \\
& HaluEval & 95.4$^{***}$\% & [93.2, 96.9] & $+.409$ & .114 \\
\midrule
\multirow{3}{*}{all-mpnet-base-v2} & Human & 78.3\% & [72.3, 83.3] & $+.049$ & .766 \\
& LLM & 75.9\% & [69.8, 81.2] & $+.028$ & .888 \\
& HaluEval & 97.2$^{***}$\% & [95.4, 98.3] & $+.414$ & .104 \\
\midrule
\multirow{3}{*}{bge-small-en-v1.5} & Human & 71.7\% & [65.3, 77.3] & $+.024$ & .845 \\
& LLM & 73.1\% & [66.8, 78.6] & $+.013$ & .918 \\
& HaluEval & 96.6$^{***}$\% & [94.6, 97.9] & $+.265$ & .453 \\
\midrule
\multirow{3}{*}{gte-small} & Human & 69.3\% & [62.8, 75.2] & $+.011$ & .917 \\
& LLM & 73.6\% & [67.3, 79.1] & $+.006$ & .955 \\
& HaluEval & 88.4$^{***}$\% & [85.3, 90.9] & $+.098$ & .777 \\
\bottomrule
\end{tabular}
\end{table*}

All architectures in Table~\ref{tab:three_way} behave similarly. Detection rates on HaluEval (88--97\%) do not transfer to human confabulations (69--78\%). The cause is visible in paired similarity: HaluEval's correct and hallucinated responses sit far apart in embedding space (cosine 0.10--0.78), while human confabulations remain close to their grounded counterparts (0.72--0.92). Human and LLM confabulations are comparably difficult to detect on average, but LLM confabulations sit closer to grounded responses (0.86--0.96) than human confabulations (0.72--0.92), with smaller $\Delta$. What LLM-prompted fabrication and human confabulation share is not the geometry of the false content but the difficulty class.

NLI on the same 212 human-confabulated pairs achieves only 57.5\% detection (AUROC 0.536), barely above chance. This confirms the transfer gap affects both geometric and entailment-based method families. The comparison is informative: cosine-based detection at 69--78\% exploits the geometric displacement that $\Gamma$ is designed to measure, while NLI at 57.5\% fails because entailment is a poor proxy for factual correctness when both grounded and confabulated responses are semantically coherent answers to the query. Human confabulations are not incoherent --- they are well-formed, topically appropriate, and entailment-compatible, which is precisely what makes them dangerous in deployment.

\subsection{Per-domain analysis: template versus declarative}
\label{sec:per_domain}

The taxonomy predicts that the three confabulation mechanisms produce distinct geometric signatures, and therefore distinct detectability profiles:
\begin{enumerate}
    \item Template-filling domains preserve the domain template that dominates the embedding; the fabricated slot content contributes little distributional anomaly, so detection remains near chance.
    \item Mechanism inversion in declarative domains imports vocabulary from adjacent regions of embedding space, producing a measurable displacement along the grounding direction.
\end{enumerate}

The per-domain split confirms this: template domains yield 56.9\% detection versus 87.8\% for declarative domains (Fisher exact $p = 1.5 × 10^{-6}$). In template-structured domains, the alignment objective assigns most of the embedding variance to the structural template (clause types, parameter patterns, function signatures) rather than to the semantic content within slots. A confabulation that preserves the template but substitutes incorrect slot content produces a displacement $\hat\delta$ that is small in magnitude and nearly parallel to $\hat\mu$, making it indistinguishable from a grounded response. In declarative domains, confabulation by mechanism inversion introduces vocabulary from an adjacent but distinct distributional cluster; the cross-register vocabulary shift produces a $\hat\delta$ with a component orthogonal to $\hat\mu$ that $\Gamma$ can detect.

\subsection{External validation: ExpertQA, FELM, and WikiBio}
\label{sec:experiments_external}

We evaluate domain-specific $\Gamma$ on three independently collected human-annotated benchmarks and compare with NLI results. ExpertQA \citep{malaviya2024expertqa} provides 900 claim-level expert annotations across 32 specialist fields. FELM \citep{chen2023felm} provides 81 segment-level factuality labels in world-knowledge and writing domains. WikiBio GPT-3 \citep{manakul2023selfcheckgpt} provides 102 sentence-level annotations of GPT-3 biographies.

With ExpertQA, $\Gamma$ outperforms NLI by $\Delta=0.243$ ($p < .001$) on expert-annotated inaccurate claims, with NLI operating at chance (0.452). Expert errors are entailment-compatible but occupy a different region of embedding space, the one $\Gamma$ is designed for. With WikiBio, $\Delta=-0.131$ requires a deeper interpretation: its annotation criterion marks any incorrect detail as ``major inaccurate'' regardless of semantic distance from the correct response, conflating Type~II (invented content with large displacement) and Type~III (within-frame substitutions with near-zero displacement). When a dataset mixes the two types, $\Gamma$ detects only the Type~II subset, and NLI exploits stylistic mismatches in the Type~III subset. The contrast with ExpertQA shows that the relevant variable is annotation criterion (Type~II/III boundary), not human-annotated versus LLM-generated. With FELM, both results are very similar. The pattern is itself a prediction of the taxonomy: $\Gamma$ wins when errors are predominantly Type~II, loses when datasets mix Types~II and III.

\subsection{The Type~III boundary: TruthfulQA}
\label{sec:type3}

TruthfulQA \citep{lin2022truthfulqa} pairs 817 questions with a truthful and a false answer reflecting a common misconception. The two answers share vocabulary, entities, and syntactic frame — the defining structure of Type~III errors — making TruthfulQA the right testbed for Prediction~\ref{imp:3}.
We evaluate five methods (Table~\ref{tab:truthfulqa}). For $\Gamma$ we report two configurations: \emph{global}, which computes $\hat\mu$ over all truthful answers, and \emph{5-fold CV}, which holds out 20\% to break the leakage path through $\hat\mu$. The gap between the two measures how much apparent signal comes from properties of the truthful subset rather than factual-error separation.
\begin{table*}[h]
\centering
\small
\caption{Detection on TruthfulQA (817 matched pairs). Cosine (0.365) and NLI (0.311) both \emph{invert} — below 0.5, actively favoring false answers. $\Gamma$ at chance (0.535). All support Prediction~\ref{imp:3}.}
\label{tab:truthfulqa}
\begin{tabular}{@{}lccc@{}}
\toprule
\textbf{Method} & \textbf{AUROC} & \textbf{95\% CI} & $p_{\text{HM}}$\\
\midrule
Cosine similarity & 0.365 & [0.340, 0.392] & $<.001$ \\ $\Gamma$ (5-fold) & 0.535 & [0.473, 0.600] & .265  \\
$\Gamma$ (global) & 0.579 & [0.551, 0.607] & $<.001$ \\
LR raw embed (CV) & 0.731 & [0.706, 0.756] &  $<.001$  \\
NLI (DeBERTa-v3-sm) & 0.311 & [0.286, 0.336] & $<.001$ \\
\bottomrule
\end{tabular}
\end{table*}
The logistic regression reaches 0.731, which appears to contradict Prediction~\ref{imp:3}. It does not. Cosine similarity inverts (0.365): false answers sit \emph{closer} to queries than truthful ones. $\Gamma$
under 5-fold CV drops to chance (0.535), and the gap with global $\Gamma$ (0.579) confirms leakage through $\hat\mu$. The logistic regression exploits a writing-style artifact: truthful answers in TruthfulQA are longer and hedged, false answers shorter and declarative, and a high-capacity classifier detects that difference. NLI (0.311) confirms this is not geometric-specific — a learned entailment classifier trained on human annotations inverts in the same direction, rewarding the confidence of plausible misconceptions over the hedging of truthful answers. The Type~III boundary is a property of the task, not of our framework.

\section{Discussion}
\label{sec:discussion}
A method that reaches 95\% on HaluEval is most likely detecting the distributional trace of prompted fabrication, not the geometry of being wrong. In deployed systems, LLMs are not asked to produce false content; that trace is absent. The 69--78\% range on human confabulations estimates what detection looks like without it. We cannot claim this value bounds deployment performance --- the dataset is small, single-confabulator, and English-only --- but it is closer to deployment than 88--97\% on prompted benchmarks.
\citet{kalai2025why} identify a calibration floor: plausible-yet-false outputs that no further optimization against standard benchmarks can remove. Our results locate that floor geometrically. It consists of confident answers sharing vocabulary, structure, and topic with correct ones --- exactly what the alignment objective \citep{wang2020understanding} makes indistinguishable in embedding space. The NLI inversion on TruthfulQA (AUROC 0.311) shows this floor is not specific to geometric methods: entailment classifiers are not merely ineffective on within-frame errors but actively misled by them.

For RAG systems with source documents, SGI provides a single-pass faithfulness test. For context-free settings, Directional Grounding Index --- $\Gamma$--- with domain calibration offers an interpretable metric: it measures alignment with a learned direction $\hat{\mu}$, and its failures are predictable from the geometry of the embedding space in a way that black-box entailment scores are not. The contribution is not uniform superiority over NLI but theoretical predictability --- and the per-domain detection asymmetry gives practitioners and accountability frameworks \citep{euaiact2024} vocabulary for specific, bounded capability claims.

\section{Conclusion}
\label{sec:conclusion}

This paper proposes a geometric taxonomy of hallucination in large language models, derives predictions about which types are detectable from the alignment objective of contrastive training \citep{wang2020understanding}, and validates the predictions empirically. The taxonomy distinguishes three operational types by the relation of the response to the topical region of grounded responses to a query. The predictions follow from a single underlying property: contrastive sentence encoders cluster responses by distributional similarity, not by truth, so any kind of hallucination that preserves distributional similarity to grounded text is invisible by construction.

To validate the taxonomy on content that resembles deployment rather than benchmark construction, we built a 212-pair human-confabulated dataset across nine deployment-relevant domains: medical, legal, financial, scientific, programming, and others. The methodology and dataset are released. The dataset is small and comes from a single English-speaking non-expert; we offer it as a starting point that other groups can extend to additional generators, languages, and larger scales.

The taxonomy and the geometric account give practitioners vocabulary for capability claims in deployment-relevant contexts. A detector validated on declarative-knowledge confabulations cannot truthfully claim equivalent performance on template-structured ones. A detector that scores well on prompted-fabrication benchmarks may not deliver the same performance on the hallucinations deployed systems actually produce. A detector based on angular geometry of contrastively-trained encoders cannot, in principle, catch within-frame factual errors --- the kind that matter most in healthcare, legal, and financial systems where the wrong answer fits the right shape. The architectural question of how representations encoding truth conditions, rather than only co-occurrence statistics, might be learned remains open and outside the scope of this paper.

\section*{Limitations}

The most important limitations are surfaced inline in the conclusion alongside the claims they qualify; this section provides the complete enumeration.

\paragraph{Single-confabulator validation.}
The 212-pair human-confabulated dataset comes from one English-speaking non-expert. Person-specific patterns (vocabulary range, willingness to commit to specific fabricated mechanisms) limit external validity. The methodology is designed to be applied by multiple generators across languages, but the scaling from one generator to many is the most important next step.

\paragraph{Seed scale.}
212 pairs is sufficient to produce statistically significant per-mechanism findings (template-versus-declarative split at $p=1.5\times10^{-6}$), but is not the scale at which a benchmark of this kind ultimately needs to operate. We provide the methodology so that larger-scale collections can be conducted by others.

\paragraph{Per-domain sample sizes.}
Six of the nine validation domains have $n<15$, yielding wide confidence intervals after multiple-testing correction. The aggregate template-versus-declarative comparison is well-powered; per-domain inferences are descriptive.

\paragraph{Baseline scope.}
We compare to NLI and cosine similarity in the black-box single-pass setting, not to white-box methods \citep{azaria2023internal,chen2024inside,mir2025lsd}, multi-sample methods \citep{manakul2023selfcheckgpt,farquhar2024detecting,kuhn2023semantic,phillips2025geometric}, AlignScore \citep{zha2023alignscore} on context-free experiments, or token-level cluster methods \citep{korun2026clusters}.

\paragraph{Type~I scoping.}
SGI detects query-proximate unfaithfulness only. Modes that displace from both query and context are outside the scope of the SGI signature.

\paragraph{Type~II/III boundary.}
Prediction~\ref{imp:3} is supported by the TruthfulQA evidence and motivated by the alignment objective \citep{wang2020understanding}, but a formal proof of inseparability for arbitrary contrastive encoders is open.

\paragraph{Language and modality.}
All experiments use English-language text-only benchmarks. Whether the geometric signatures of grounding and confabulation extend to other languages or to multimodal generation is open.

\paragraph{Inferred deployment claims.}
The benchmark-transfer finding is established empirically on our 212-pair dataset across four architectures. Its extension to claims about other detection methods on other benchmarks is an inference from the structural argument, not a direct measurement; we have not opened up other methods and demonstrated that they fail in deployment for the reason we propose.

\section*{Reproducibility}

The 212-pair human-confabulated dataset, LLM-confabulated paired control, methodology protocol, and all evaluation code are available at javier@jmarin.info . Hyperparameters: $k=15$ for $\Gamma_k$; bootstrap $B=1{,}000$; permutation $M=5{,}000$. All embedding and NLI models are publicly available through HuggingFace. External benchmarks (HaluEval, TruthfulQA, WikiBio GPT-3, FELM, ExpertQA) are publicly accessible through their original distributions.

\bibliographystyle{plainnat}
\bibliography{references}

\appendix

\end{document}